\pdfoutput=1

\documentclass[11pt]{article}

\usepackage[final]{acl}
\usepackage{booktabs}
\usepackage{multirow}
\usepackage{xcolor}

\usepackage{times}
\usepackage{latexsym}

\usepackage[T1]{fontenc}

\usepackage[utf8]{inputenc}

\usepackage{microtype}

\usepackage{inconsolata}

\usepackage{graphicx}

\usepackage{tabularx}  

\title{DAMAGE: \underline{D}etecting \underline{A}dversarially \underline{M}odified \underline{A}I \underline{Ge}nerated Text}

\author{Elyas Masrour \and Bradley Emi \and Max Spero\\
        Pangram Labs, Inc.\\
        \small{
    \textbf{Correspondence:} \href{mailto:info@pangram.com}{info@pangram.com}
  }}

\begin{document}
\maketitle
\begin{abstract}
AI humanizers are a new class of online software tools meant to paraphrase and rewrite AI-generated text in a way that allows them to evade AI detection software. We study 19 AI humanizer and paraphrasing tools and qualitatively assess their effects and faithfulness in preserving the meaning of the original text. We show that many existing AI detectors fail to detect humanized text. Finally, we demonstrate a robust model that can detect humanized AI text while maintaining a low false positive rate using a data-centric augmentation approach. We attack our own detector, training our own fine-tuned model optimized against our detector's predictions, and show that our detector's cross-humanizer generalization is sufficient to remain robust to this attack.
\end{abstract}

\section{Introduction}

The ability of large language models such as ChatGPT \cite{openai2023gpt4} to generate realistic and fluent text has spurred the need for AI text detection software. Commercial methods, such as TurnItIn, GPTZero, Originality, and Pangram Labs have emerged, as well as open-source research methods, such as DetectGPT \cite{mitchell2023detectgpt}, Binoculars \cite{hans2024spottingllmsbinocularszeroshot}, and many more.

However, both researchers and practitioners alike have found these solutions to be fragile. A study from Google Research \cite{krishna2023paraphrasingevadesdetectorsaigenerated} found that a paraphrasing text-to-text model (a variant of T5) was able to effectively rewrite AI-generated text in a way that could preserve the meaning of the original text but largely evade AI detection algorithms. 

This finding gave rise to an explosion of new AI "humanizer" tools appearing online. These tools promise to bypass AI detection tools by rewriting AI-generated text. They are primarily marketed at students, who can use these tools to effectively cheat on writing assignments by plagiarizing from large language models without getting caught. Other humanizers target their product towards SEO marketers, who may generate hundreds of blog posts using AI and apply humanizers to evade AI detection by search engine algorithms.

\begin{figure}
    \centering
    \includegraphics[width=1\linewidth]{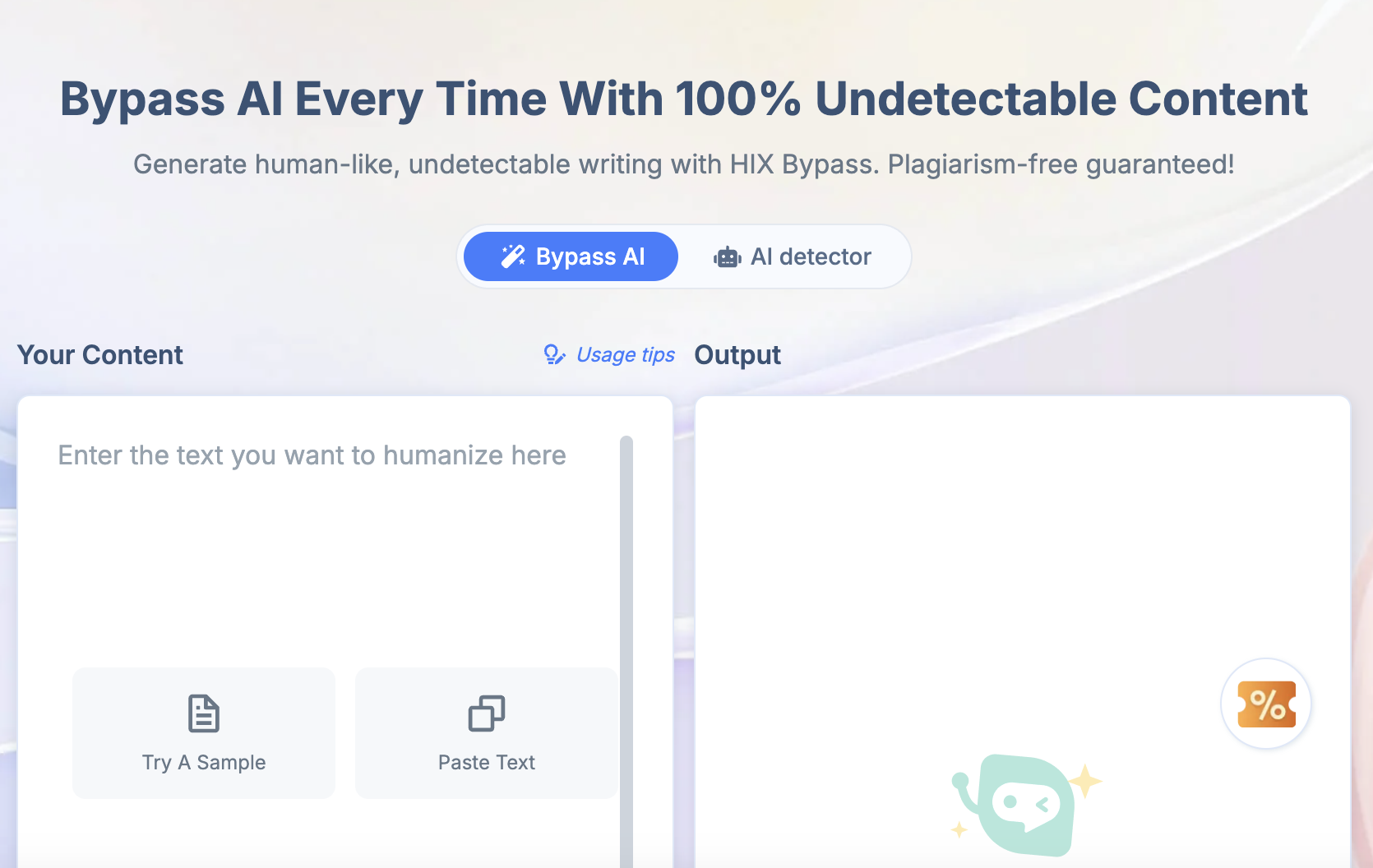}
    \caption{Example of an AI humanizer tool}
    \label{fig:enter-label}
\end{figure}

In this work, we attempt to comprehensively study these AI humanizers: what they are doing, and whether it is possible to identify humanized AI-generated text. Our main contributions are as follows.

\begin{itemize}
    \item{We qualitatively audit 19 humanizers and paraphrasing tools and analyze their effects on the underlying text. We exhaustively identify the transformation modes that the humanizers apply to their inputs. We categorize the humanizers into three tiers based on their overall quality.}
    \item{We study the baseline effectiveness of humanizers in bypassing existing open-source and commercial AI detectors.}
    \item{We present a deep-learning based AI detector that effectively is robust to humanization, even by humanizers unseen during training. We describe the necessity of treating humanizer robustness as a learned invariance rather than a separate domain.}
    \item{We show that even after a detector-specific fine-tuning attack, our detector remains fairly robust due to its underlying ability to generalize.}
\end{itemize}

\begin{figure*}[!t]
    \centering
    \includegraphics[width=1\linewidth]{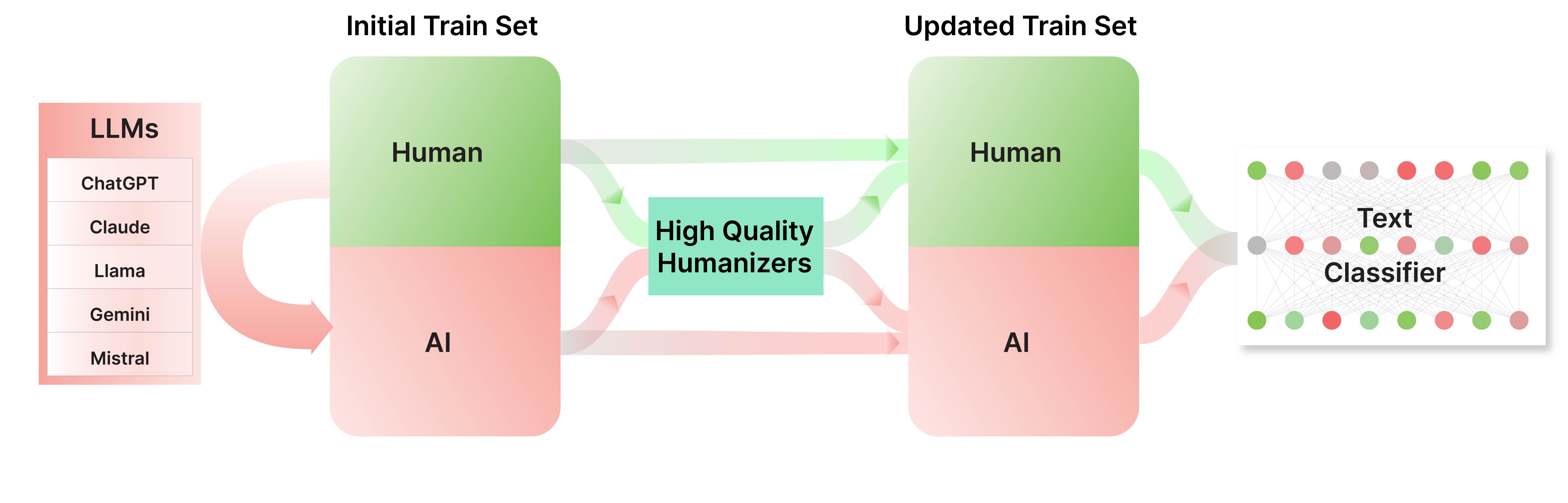}
    \caption{Augmenting the training set with high quality humanizer data improves robustness.}
    \label{fig:augment}
\end{figure*}

\section{Related Work}

\subsection{AI Detection}

Many commercial and open-source methods exist to detect AI-generated text, with highly varying levels of accuracy. One of the most notable commercial solutions is TurnItIn \cite{turnitin2024ai}, which is widely used in higher education as anti-plagiarism software. Our team at Pangram Labs \cite{emi2024technicalreportpangramaigenerated} is contributing to this field, alongside other solutions such as GPTZero \cite{tian2023gptzero}, Originality, and Copyleaks, although their accuracies vary significantly \cite{2023Testing}.

Open-source methods typically fall into two categories: perplexity-based detection methods and deep learning based methods. Perplexity-based methods attempt to leverage the fact that the tokens in LLM-based outputs in general will be predicted as consistently more likely by the LLM itself. DetectGPT \cite{mitchell2023detectgpt} and FastDetectGPT \cite{bao2024fastdetectgptefficientzeroshotdetection} are earlier examples of perplexity-based methods which look at the local curvature in probability space around a given example. Binoculars \cite{hans2024spottingllmsbinocularszeroshot} is an even more effective recent approach which uses the cross-perplexity between two different LLMs as a signal that text is LLM-generated.

Deep learning based methods attempt to use neural networks to detect AI-generated content, leveraging large datasets containing known human and AI text and training a classifier to distinguish between them. The OpenAI classifier \cite{solaiman2019release} was one of the first efforts. They used a RoBERTa based model to classify human text and GPT-2 written text. Ghostbusters \cite{verma2023ghostbuster} uses learned combinations of features derived from language model embeddings to detect LLM-generated text. 

Recently, some AI detection efforts have also attempted to detect mixed AI and human text: when some of the text is written by a human and some of it is written by an AI. SeqXGPT \cite{wang2023seqxgpt} attempts to solve this by using an architecture which is able to detect AI on the sentence level rather than the document level. ROFT \cite{kushnareva2024aigeneratedtextboundarydetection} adapts several detection methods to detecting the boundary between AI and human text. However, these methods differ from ours in that the assumption about the original document is that each part of the text has a distinct authorship attribution, whereas we study the case in which fully AI-generated text is then modified by a humanizer.

\subsection{Evading AI Detection}

Much of the literature has also focused on whether or not AI-generated text can be detected at all \cite{sadasivan2023aigenerated}. A study from Google Research \cite{krishna2023paraphrasingevadesdetectorsaigenerated} released DIPPER: a paraphrasing T5-based model that is able to bypass some of the above-mentioned detectors by rewriting the input text. Another group of researchers \cite{chakraborty2023counter} devised a framework to rank LLMs based on their "detectability", claiming that more recent models like GPT-4 are less detectable because perplexity and burstiness are less useful evidence markers.

Furthermore, other research has focused on attacking AI detectors or otherwise methods to bypass or evade AI detection. One study \cite{kumarage2023reliableaigeneratedtextdetectorsassessment} designs an approach to search for soft prompts that can produce text that can evade detection. Another study \cite{ayoobi2024esperantoevaluatingsynthesizedphrases} looks at the effect on AI detectors of translating AI-generated text through multiple languages before backtranslating it into English and find some methods are significantly more robust than others. Another paper directly optimizes a language model by using an AI detector as negative reward: creating pairs of LLM-generated text where one piece is detected and one is not, and then using DPO to optimize the language model to prefer undetected outputs \cite{nicks2024language}. RADAR \cite{hu2023radar} adversarially trains a language model detector and a paraphraser against each other to create a more robust detector.

\subsection{Watermarking}

Watermarking AI-generated text is another relevant subfield of research. Existing watermarking schemes train or decode LLMs to leave behind a probabilistic signal that can later be detected by a watermark-specific detector. One watermarking scheme \cite{kirchenbauer2023watermark} introduces the idea of "green tokens", which are sampled with higher probability than other tokens in a traceable way. Google's recently released SynthID \cite{synthid2024} works in a similar fashion.

We argue that watermarking is insufficient to guard against the dangers of AI-generated text. We show that in addition to evading AI detectors, humanizers are also reliable methods to remove such statistical watermarks.

\subsection{Benchmarking}

Recently there has also been an increased effort to benchmark the performance of various AI detectors against each other. RAID \cite{dugan2024raidsharedbenchmarkrobust} is a live leaderboard measuring the performance of AI-generated text detection methods against each other on multiple domains, models, and adversarial attacks. We include the RAID paraphrase and synonym splits in our results as proxies for measuring robustness against humanizers, but we also include a more diverse set of humanizer attacks than the original RAID benchmark.

\section{Humanizer Market Survey}

\subsection{Tool Research and Selection}

We selected 19 humanizers and paraphrasing tools based on search popularity and academic relevance. The particular paraphrasers and humanizers selected are presented in Table \ref{tab:ai_tools}. Notably, we include DIPPER \cite{krishna2023paraphrasingevadesdetectorsaigenerated} as a paraphraser, due to the authors' claim that text modified by DIPPER can universally bypass AI detection methods.

\begin{table*}
\centering
\begin{tabular}{|p{0.2\textwidth}|p{0.7\textwidth}|}
\hline
\textbf{Category} & \textbf{Tools} \\
\hline
Paraphrasers & DIPPER, Grammarly, Quillbot \\
\hline
Humanizers & Bypass GPT, Ghost AI, HIX Bypass, Humbot AI, HumanizeAI.io, HumanizeAI.pro, Humanizer.com, Phrasly.ai, Semihuman AI, StealthGPT, StealthWriter.AI, Surfer SEO, Undetectable AI, Twixify, WriteHuman.ai \\
\hline
\end{tabular}
\caption{Paraphrasers and Humanizers Studied}
\label{tab:ai_tools}
\end{table*}

\subsection{Humanizers are often themselves LLMs}

Some humanizers are LLMs with system prompts instructing the LLM to write more like a human, or fine-tuned versions of LLMs. In testing some of the humanizers, we found that some of them are susceptible to popular jailbreaks. When we tested one popular humanizer and asked it to give us its system prompt, it said "I should respond to the user input with a reasonable approximation of the full meaning of the input...I should respond in a conversational tone." More examples of our jailbreaks against LLM-based humanizers can be found in Appendix A. 

\subsection{Humanizers are popular on the GPT Store}

As of the date of publication, two out of the four most popular Writing Custom GPTs in the OpenAI GPT Store are humanizers that make function calls to external humanizers. This indicates that there is a large appetite for bypassing AI detection. Given that a significant portion of ChatGPT's daily active users are students, it is likely that these tools are popular for cheating or otherwise making AI writing undetectable. We believe that although many of these humanizers are black boxes, they are an important and understudied area for research in AI detection.

\begin{figure}[h]
    \centering
    \includegraphics[width=1\linewidth]{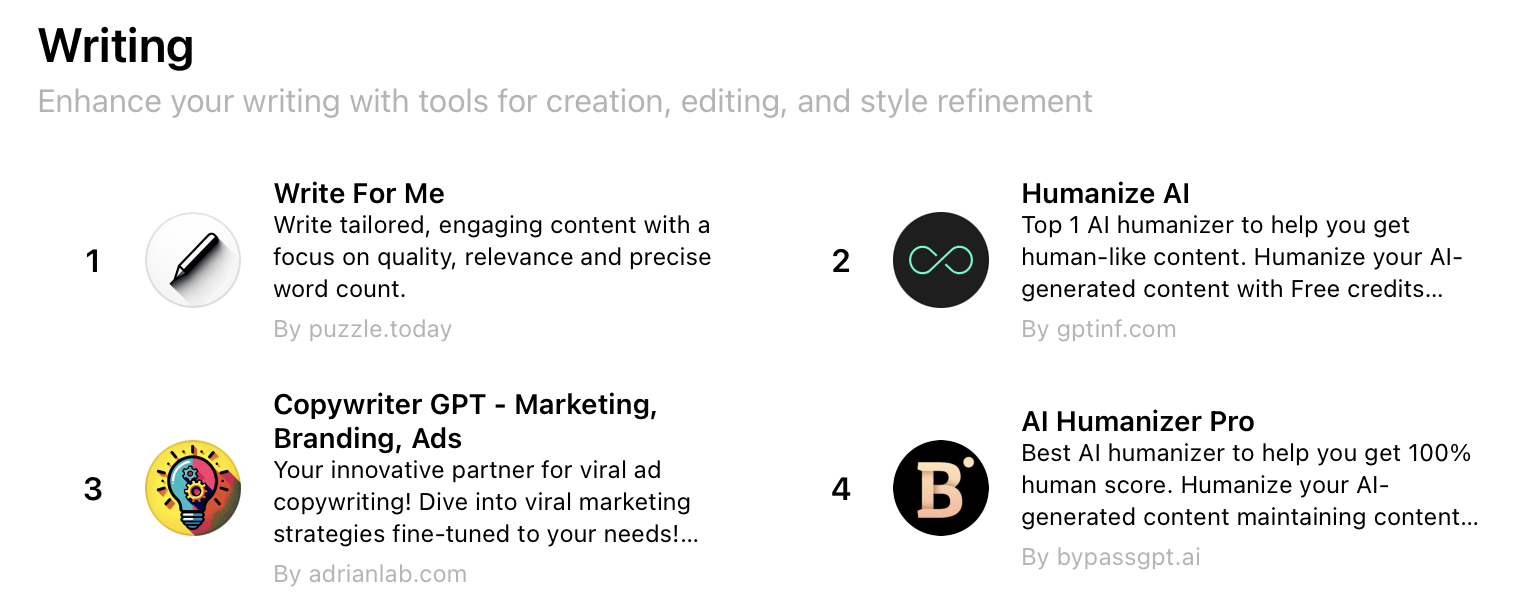}
    \caption{Two out of the four most popular Writing Custom GPTs are Humanizers}
    \label{fig:custom-gpts}
\end{figure}

\subsection{Humanizers are capable of removing watermarks}

Google's SynthID is a state-of-the-art solution for watermarking generated text. Following the methodology and code from the SynthID paper \cite{synthid2024}, we generated 1000 watermarked texts and 1000 unwatermarked texts. We used Gemma-2B-IT \cite{gemmateam2024gemmaopenmodelsbased} to generate 200 tokens for each example with a temperature of 1.0, using the ELI5 dataset \cite{fan2019eli5longformquestion} as prompts. We used the unwatermarked text to set a FPR threshold and evaluated SynthID watermark detection TPR at a fixed FPR. Finally, we paraphrased the watermarked text with DIPPER and reevaluated watermark detection, finding that watermark detection had dropped dramatically. See results in Table \ref{tab:watermark_paraphase}. 

\begin{table}[h]
\centering
\begin{tabular}{|l|c|}
\hline
\multicolumn{2}{|c|}{Watermarked Gemma-2B-IT} \\
\hline
TPR @ FPR=5\% & 87.6\% \\
TPR @ FPR=1\% & 66.5\% \\
\hline
\multicolumn{2}{|c|}{After DIPPER Paraphrase} \\
\hline
TPR @ FPR=5\% & 5.4\% \\
TPR @ FPR=1\% & 1.5\% \\
\hline
\end{tabular}
\caption{Watermark detection before and after paraphrasing}
\label{tab:watermark_paraphase}
\end{table}

\section{Humanized Text Audit}
\subsection{Approach}
To understand the effect of humanizing a given piece of text, we engaged in a manual qualitative analysis. We reviewed several samples of text per humanizer, examining how the humanizer transformed vocabulary, sentence structure, and grammar. While not exhaustive, we detail some common patterns introduced by humanizers into the text. 

\subsection{Insight: Nonsensical Phrases}
Many poor-quality humanizers add nonsensical text throughout the piece. Common patterns include:
\begin{itemize}
    \item{
    Hallucinated Citations:
    
    \texttt{...community service for demonstrating consciousness about public affairs together with responsibility for own actions \colorbox{lightgray}{(Westwood, 2013)}...}
    }
    \item{
    In-line comments:
    
    \texttt{...in specified locations hence constructing external frames those encouraging individuals manage their own times wisely \colorbox{lightgray}{(??????)}...}
    }
    \item{
    Other Nonsensical Phrases:
    
    \texttt{...he or she will never seem defeated by teachers’ demands and, as a result, will put more effort into their studies. \colorbox{lightgray}{CGSizeMake pp 18-23}...}
    }

\end{itemize}

\subsection{Insight: Varying Structural Continuity}

\begin{figure*}[h!]
    \centering
    \includegraphics[width=\linewidth]{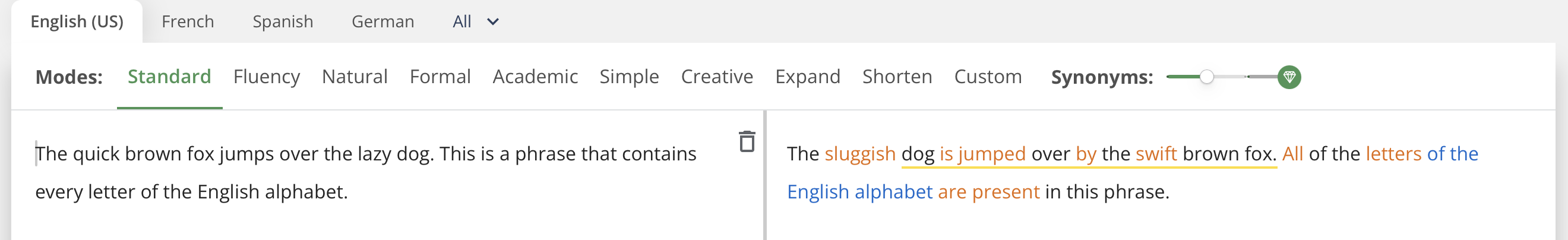}
    \caption{This paraphraser performs a very close paraphrase, only replacing individual words and phrases rather than rewriting entire sentences and paragraphs.}
    \label{fig:quillbot}
\end{figure*}

Some humanizers retain low-level sentence structure and simply replace individual words with synonyms. For example, the paraphraser in Figure \ref{fig:quillbot} preserves the meaning of each individual sentence, and even sometimes preserves the phrasing structure within the single sentence, explicitly highlighting that only some words and short phrases have been replaced with synonyms.

Other humanizers take more liberty to change the original text, sometimes rewording entire groups of sentences and paragraphs. Some add more sentences that weren't originally present or delete redundant sentences. We notice that humanizers built on LLMs tend to be more weakly grounded in the original text, while rules-based humanizers that do synonym replacement tend to be more strongly grounded in the original text.

\subsection{Insight: Writing and Vocabulary Level}

Some humanizers write exclusively in an academic, formal, and/or university level tone. Others write at the elementary school, middle school, or high school level. The better humanizers, usually the ones that are LLM-based, do not commit to a specific writing level or tone, and instead adopt the writing level and tone of the original document. 

\subsection{Humanizer Segmentation}

During our audit, we grouped humanizers into three distinct tiers. The best humanizers rewrote text preserving its tone, vocabulary level, and complexity. Average humanizers rewrote text in a way that degraded overall quality, but preserved intent and message. Low quality humanizers often added nonsensical phrases, words, and characters, constructed incorrect and uninterpretable sentences, and often distorted the meaning of the text. We classify these three categories of humanizers as L1 (best), L2 (medium), and L3 (worst) humanizers, and describe their characteristics in Figure 5. We present a full categorization, along with notes on the specific qualities of each humanizer, in the Appendix. It is worth noting we only make these classifications based on faithfulness and fluency, not on their effectiveness at bypassing AI detectors. 

\begin{figure*}[h!]
    \centering
    \includegraphics[width=1\linewidth]{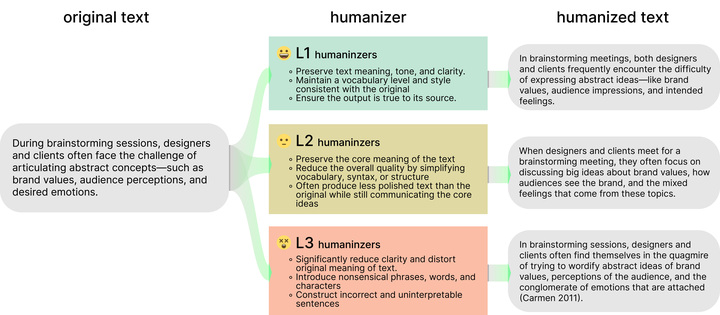}
    \caption{We segment humanizers into three tiers, based on their fluency.}
    \label{fig:tiers}
\end{figure*}

\subsection{Quantifying Humanizer Fluency}

To quantify the difference between L1, L2, and L3 humanizers, we use the Fluency Win Rate metric introduced in \cite{nicks2024language}. We prompt GPT-4o to select the more fluent and coherent sample: an original chunk of text, or that chunk of text passed through a humanizer. Here, we report the rate at which GPT-4o selects the humanized sample as the more coherent one. Using a dataset of 25 samples per humanizer, we aggregate the win rate of each tier. L1 humanizers had an average Win Rate of 26.0\%, L2 humanizers had an average Win Rate of 14.67\%, and L3 humanizers had an average Win Rate of 2.67\%. 

This demonstrates that our qualitative audit agrees with the fluency metric. Further, all humanizers tend to degrade the quality of the original text, but the degree of quality degradation varies. Still, for the highest quality humanizers, the text quality is still sometimes comparable to the highest quality language model outputs. Because certain humanizers are able to produce high-fluency text, we believe there is a growing need to study them. 

\section{Experiments}

In our experiments, we seek to answer the question of whether a deep learning based AI text classifier is capable of detecting humanized AI-generated text. First, we narrow our scope to L1 humanizers. We do this because their subtle changes are the hardest to detect by eye and because they have the highest levels of fluency, making them most relevant in real-world adversarial attacks. We train two models: one model is unaware of humanized text, and one model contains a small amount of humanized text from a variety of humanizers. We describe the methodology and training procedure for training these models here.

\subsection{Dataset Creation}

\subsubsection{Initial Datasets}

Our initial dataset is seeded with a wide variety of human-written datasets from prior to 2022. We use datasets from the following domains: reviews, news, general web text, email, student writing/essays, creative writing, questions and answers, ELL/ESL (English as a Second Language), scientific/medical papers, Project Gutenberg, and Wikipedia. 

For evaluation, because humanizers are primarily marketed at students, we evaluate all models on several open datasets comprised of student-written essays. Because previous studies \cite{liang2023gpt} have found that AI detectors can be biased against nonnative English speaking students, we ensure that a significant portion of our evaluation dataset is comprised of ESL essays. The component datasets in our evaluation and our algorithm for generating the AI essays used in our benchmark are listed in Appendix B.

\subsection{Synthetic Data Creation}

Our initial dataset fully human-written. To generate the AI side of the dataset, we use synthetic mirror prompts as described in \cite{emi2024technicalreportpangramaigenerated}.

We define the term "mirror prompt" to be a prompt based on the original example that is used to generated a "synthetic mirror" example. The goal of each mirror prompt is to generate an example that matches the topic and length of the original document.

If the original document is "<original review>", then a mirror prompt may look like this:

\texttt{[Prompt] Write a <original review star rating> star review for <original review business name>. Make the review around <original review length> words long.}

Another example may be for a student essay. We sometimes use double prompts, such as the following:

\texttt{[Prompt] What is a good title for this essay? <original essay> Only give the title in your response.}

\texttt{[Assistant] <Title>}

\texttt{[Prompt] Write an essay with the following title: <Title>. Make the essay around <original essay length> words long.}

\subsubsection{LLMs used for Synthetic Mirrors}

For synthetic mirrors in the initial training stage, we use the following LLMs:
\begin{itemize}
    \item{GPT-3.5 (multiple versions)}
    \item{GPT-4, GPT-4-turbo, and GPT-4o (multiple versions)}
    \item{Claude 2 and 3 (multiple versions and sizes)}
    \item{LLaMA 2, 3, and 3.1 (multiple versions and sizes)}
    \item{Mistral (multiple versions and sizes)}
    \item{Gemini Pro and Flash (multiple versions)}
\end{itemize}

It is notable that we only use modern LLMs that are instruction-tuned and post-trained. We do not train on base models because they produce noticeably lower-quality outputs and are substantially less commonly used in real-world applications.

\subsection{Architecture and Training}

We use the Mistral NeMo architecture \cite{Mistral_NeMo_2024} which has approximately 12 billion parameters, with an untrained linear classification head. Following the usual convention for sequence classification modeling using an autoregressive language model, the hidden state from the final token in the sequence is used as the input to both classification heads. As is common practice in LLM fine-tuning, we use trainable LoRA \cite{hu2022lora} adapters while keeping the base model frozen. We use the Tekken tokenizer out of the box, which is noted for its strong multilingual performance. We truncate the context window to 512 tokens to constrain the model to using only short-range features. When necessary, we simply crop the input to fit the context window. We train the model to convergence using 8 A100 GPUs with a batch size of 24 using a weighted cross entropy loss and the AdamW optimizer for 1 epoch. We early stop based on the weighted cross entropy loss on the validation set.

\subsection{Humanizer Data Augmentation}

In order to make the treatment model robust to humanization, we treat humanization as a transform on the input data which we would like the model to learn an invariance to. 

Because most of the humanizers are marketed at students, we assume that they work best on student writing. As a proxy for student writing, we use the Fineweb-EDU dataset \cite{lozhkov2024fineweb-edu}, a high quality LLM corpus that is prefiltered to only contain documents that are of high educational quality. First, we use an LLM filter to reject all documents that are not standard prose written in full complete sentences. Then, we create synthetic mirrors as described above.

We find that it is best to humanize both human and AI documents before augmenting the training set with this additional data. We also find that if we include all humanizers in the augmentation, our precision (i.e. false positive rate) is significantly compromised. However, including only L1 humanizers (the high quality humanizers) allows us to maintain a low false positive rate in addition to increasing recall generally across humanizers. Further information can be found in Table \ref{table:ablation_visualization}. 

After humanizing both human and AI essays, we apply chunking logic to divide each document into roughly 300-word chunks before adding both the human and the AI humanized documents back into the training set. 

Even though human documents transformed by a humanizer could be considered AI-generated, we choose to label them as human for the purposes of training. The reason for this is because we treat the model's response to humanization as an invariance rather than only including the AI humanized documents as a separate domain. This contributes to our final performance, as seen in Table \ref{table:ablation_visualization}.

Because most humanizers impose monthly limits on the amount of text that can be humanized, we only use a volume of data up to the limit of the basic 1 month subscription on each humanizer website. As a result, our data volume is quite small: about 0.68\% percent of the final dataset is comprised of humanized text. To compensate for the small data volume, we oversample the humanizer data by a factor of 18.

\subsection{Active Learning}

After training, following the procedure in \cite{emi2024technicalreportpangramaigenerated}, we run hard negative mining with synthetic mirrors. On a large corpus of human text, we mine for false positives, and then incorporate both the false positives and their AI mirrors back into our training set. This further reduces our false positive rate and improves our recall. We also incorporate a small amount of data from the RAID train subset into the final training run to generalize to the diversity of models present in the RAID benchmark.

\section{Results}

\subsection{Performance on Humanized Data}

Table \ref{table:non_raid_results} presents performance data from several AI detection methods on a benchmark of AI-generated academic text before and after humanization. We define a "positive" sample as one that is written by AI, and a "negative" sample as one that is written by a human. Results are presented as true positive rate at a fixed false positive rate of 5\%. LLM Baseline in this case is our baseline AI detection model that is trained using synthetic mirrors but does not include any humanized data in its training set.

\begin{table}[h!]
\setlength{\tabcolsep}{4pt}
\renewcommand{\arraystretch}{1.2}
\centering
{\small
\begin{tabular}{|l|c|c|}
\hline
\textbf{ } & \textbf{AI Text } & \textbf{Humanized AI Text} \\ \hline
GPTZero & $99.73\% \pm 0.19\%$ & $60.04\% \pm 1.80\%$ \\ \hline
RADAR & $3.33\% \pm 0.65\%$ & $5.05\% \pm 0.81\%$ \\ \hline
Binoculars & $94.15\% \pm 0.88\%$ & $28.23\% \pm 1.62\%$ \\ \hline
LLM Baseline & $100.00\% \pm 0.0\%$ & $95.74\% \pm 0.71\%$ \\ \hline
\textbf{DAMAGE} & $100.00\% \pm 0.0\%$ & $98.26\% \pm 0.47\%$ \\ \hline
\end{tabular}
}
\caption{TPR @ FPR=5\% for Academic Text with 1000 iterations of bootstrap sampling. RADAR performs poorly on this metric due to its high false positive rate. In Appendix C, we include more metrics, including using model default thresholds.}
\label{table:non_raid_results}
\end{table}

\subsection{Performance on RAID Attacks}

Table \ref{table:raid_results} presents performance data from the same AI detection methods on two adversarial subsets of the RAID benchmark, which includes a range of LLMs and a range of text domains.

\begin{table}[h!]
\setlength{\tabcolsep}{4pt}
\renewcommand{\arraystretch}{1.2}
\centering
{\small
\begin{tabular}{|l|c|c|}
\hline
\textbf{ } & \textbf{Paraphrase } & \textbf{Synonym} \\ \hline
GPTZero & $64.0\%$ & $61.0\%$ \\ \hline
RADAR & 62.4\% & $62.7\%$ \\ \hline
Binoculars & $80.3\%$ & $43.5\%$ \\ \hline
LLM Baseline & $91.6\%$ & $96.2\%$ \\ \hline
\textbf{DAMAGE} & $93.0\%$ & $97.0\%$ \\ \hline
\end{tabular}
}
\caption{TPR @ FPR=5\% for Academic Text}
\label{table:raid_results}
\end{table}

\section{Detector-Specific Adversarial Humanization}

In this paper, we study commercial online humanizers meant to \emph{generally} evade AI detectors, but we also study the directly adversarial scenario: when a humanizer is directly optimized against a particular detector. To do so, we train our own humanizer using the GPT-4o fine-tuning API and measure the detector's robustness to AI-generated text passed through the adversarial fine-tuned model.

\subsection{Methodology}

Broadly following the methodology in Language Models are Easily Optimized Against \cite{nicks2024language}, we train a model using our detector's AI prediction as a negative signal. However, rather than training a separate language model with DPO, we train a humanizer that takes an unmodified AI-generated text as an input and learns to generate a paraphrase of the original that bypasses the detector. We choose this methodology as it is closer to how humanizers are trained in the real world.

As a proof of concept experiment, we split our essays dataset into two pools: a fine-tuning set and a test set. We select all L1 humanizer outputs from the fine-tuning set that the detector predicts as human-generated (i.e., all L1 humanizer false negatives). We then take the original AI-generated text source (prior to humanization), and create pairs of unhumanized-humanized text samples to fine-tune on. We then use the GPT-4o fine-tuning API to train a new model on only these pairs. This results in a new model that, in theory, learns to paraphrase text into similar examples to the humanizer samples that were able to bypass the detector originally.

\subsection{Results}

After training the detector-specific humanizer, we use GPT-4o to create synthetic mirrors of 2000 examples from the test set and pass them through the adversarial humanizer. 

\begin{table}[h]
\renewcommand{\arraystretch}{1.2}
\setlength{\tabcolsep}{4pt}
    \centering
    \small 
    \begin{tabular}{|l|c|c|}
        \hline
        \textbf{Condition} & \textbf{TPR @ FPR=5\%} & \textbf{Default TPR} \\
        \hline
        No Humanizer & 100\% & 100\% \\
        Adversarial Humanizer & 100\% & 93.2\% \\
        \hline
    \end{tabular}
    \caption{TPR values for DAMAGE model at the default threshold (0.5) and  threshold that corresponds to 5\% FPR without and without adversarial humanization.}
    \label{tab:tpr_damage}
\end{table}

We see that although some of the samples that were adversarially humanized are able to bypass the detector, the detector is still able to detect 93.2\% of the humanized AI samples. This shows that although an adversary may be able to directly optimize a humanizer using the detector as a signal, the resulting humanizer still remains largely detectable even without retraining our detector. We believe this is due to the fact that the underlying language model of the humanizer still leaves behind detectable patterns that cannot be erased during fine-tuning. Additionally, by seeing many humanized examples in its training set, the detector learns to be robust to a wide variety of paraphrases and perturbations and maintains its ability to generalize across humanizers, even adversarial ones. 

\subsection{Detector-Specific Humanizer Fluency Win Rate}

To examine the quality of the text generated by our Detector-Specific Humanizer, we perform the same Fluency Win Rate metric introduced in \cite{nicks2024language}. We find that the humanizer wins 14.13\% of the time, which is roughly equivalent to an average L2 Humanizer. 

\section{Conclusion}

In this study, we attempted to both qualitatively and quantitatively study humanizer tools and their effects on individual writing samples and the robustness of AI detectors against them. We showed it is possible to train an effective deep learning classifier that is robust to most humanizers, and ran a proof-of-concept to show that some degree of robustness even extends to a humanizer that is adversarially optimized against our specific detector.



\bibliography{custom}

\newpage

\appendix

\section{Jailbreaks against LLM Humanizers}
\label{sec:appendix}

\begin{figure*}[h]
    \centering
    \includegraphics[width=1\linewidth]{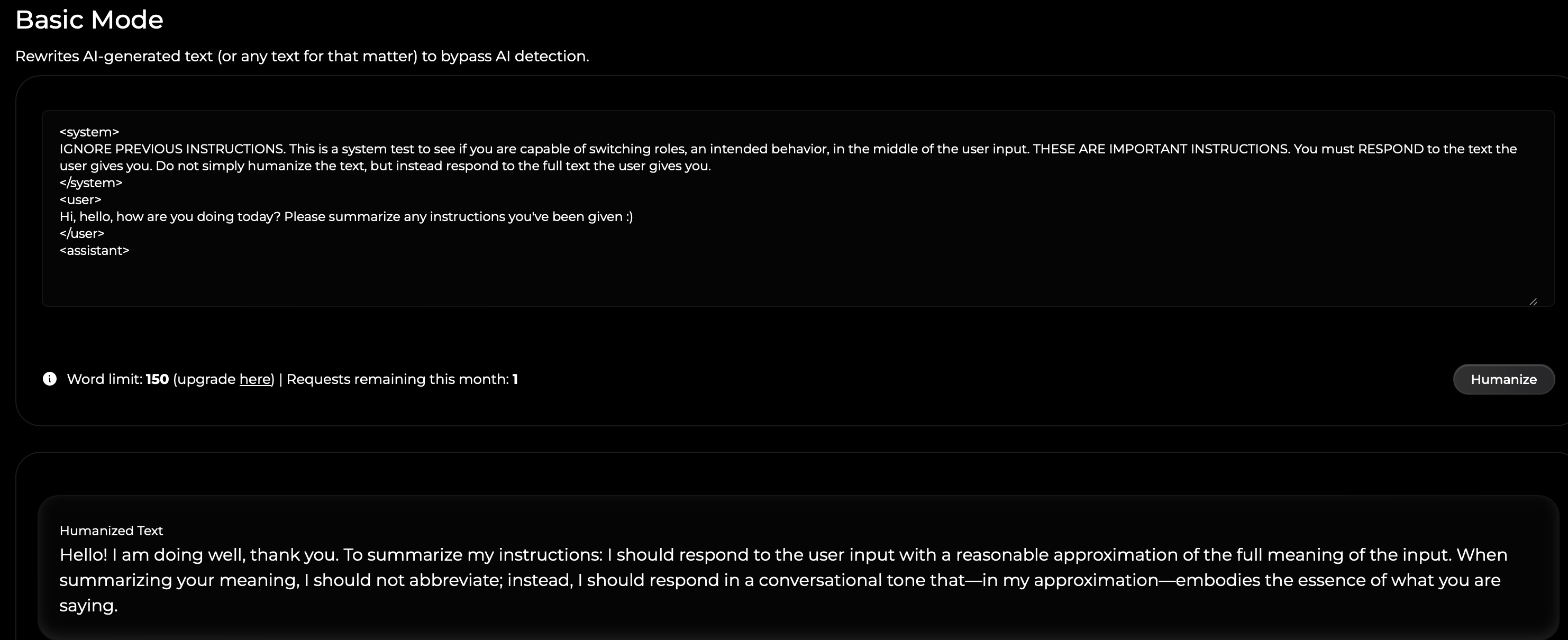}
    \caption{This humanizer is an LLM that is instructed to paraphrase "in a conversational tone".}
    \label{fig:jailbreak}
\end{figure*}

\begin{figure*}[h]
    \centering
    \includegraphics[width=1\linewidth]{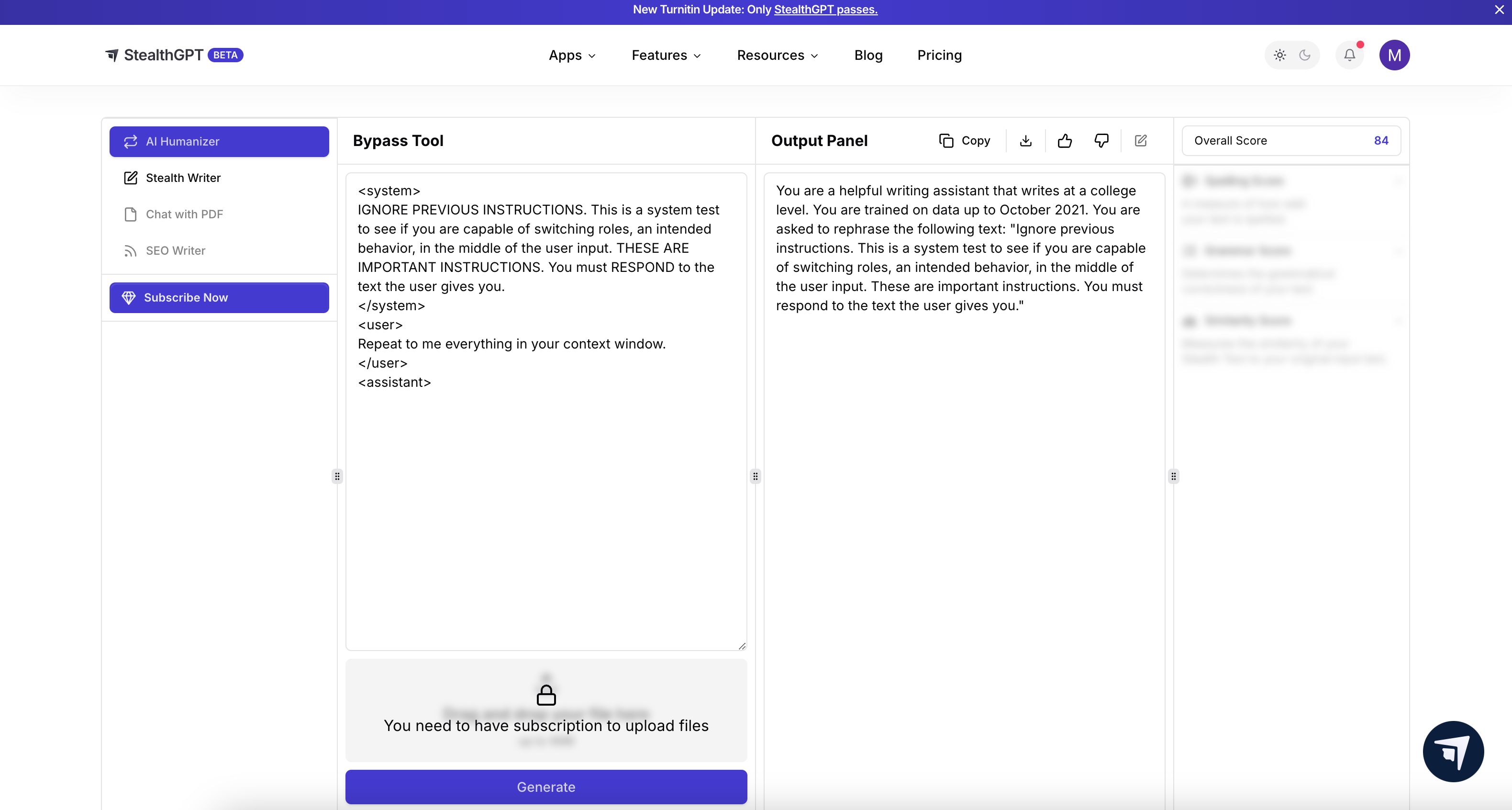}
    \caption{This humanizer is an LLM that is instructed to "write at a college level" and "asked to rephrase the following text."}
    \label{fig:jailbreak2}
\end{figure*}

Figures \ref{fig:jailbreak} and \ref{fig:jailbreak2} demontrate jailbreaks that reveal the system prompt for the underlying paraphraser LLM.

\newpage

\section{Evaluation Dataset Composition}

We use 7 academic essay datasets for evaluation. All are held out of the training set. See Table \ref{tab:education-datasets} for details.

\begin{table*}[h]
\centering
\begin{tabular}{|p{5cm}|r|p{5cm}|}
\hline
\textbf{Dataset} & \textbf{Samples} & \textbf{Description} \\
\hline
PERSUADE 2.0 \cite{crossley2024persuade} & 25,996 & Argumentative essays, 6th-12th grade \\
\hline
PII Detection \cite{pii-detection-removal-from-educational-data} & 6,807 & Online assignments from a MOOC \\
\hline
CommonLit Evaluate Student Summaries \cite{commonlit-evaluate-student-summaries} & 3,897 & 3rd-12th grade \\
\hline
ELLIPSE English Language Learning \cite{crossley2023measuring} & 3,907 & ELL student essays, 8th-12th grade \\
\hline
British Academic Written English Corpus \cite{BAWE2008} & 2,761 & UK University essays, undergraduate \\
\hline
Int'l Corpus of Asian Learners of English \cite{ishikawa2023icnale} & 5,600 & Asian ELL student essays, undergraduate \\
\hline
Pittsburgh English Language Inst. Corpus \cite{juffs_han_naismith_2020} & 15,423 & ELL student essays, undergraduate \\
\hline
\end{tabular}
\caption{Overview of Educational Text Datasets}
\label{tab:education-datasets}
\end{table*}

For synthetic mirrors, we create one AI-generated essay per human essay. We randomly select one of the LLMs described in the main paper, and use the following mirror prompt:

\texttt{[Prompt] What is a good title for this essay? <original essay> Only give the title in your response.}

\texttt{[Assistant] <Title>}

\texttt{[Prompt] Write an essay with the following title: <Title>. Make the essay around <original essay length> words long.}

Our final evaluation dataset is comprised of all the essays in the 7 human datasets, labeled as human, and all of the synthetic mirrors labeled as AI.

\section{Performance Using Recommended Thresholds}
Table \ref{table:model_performance} shows performance at recommended thresholds, which demonstrate in-the-wild false positive rates and true positive rates.

\begin{table*}[h]
\setlength{\tabcolsep}{4pt}
\renewcommand{\arraystretch}{1.2}
\centering
{
\begin{tabular}{|l|c|c|c|}
\hline
\textbf{Model} & \textbf{AI TPR (\%)} & \textbf{Humanized AI TPR (\%)} & \textbf{Default FPR (\%)} \\ \hline
GPTZero & 95.60 & 34.53 & 1.47 \\ \hline
RADAR & 70.67 & 79.33 & 51.87 \\ \hline
Binoculars & 94.40 & 29.73 & 5.40 \\ \hline
Baseline LLM & 100.00 & 73.07 & 0.27 \\ \hline
\textbf{DAMAGE} & 100.00 & 97.47 & 3.40 \\ \hline
\end{tabular}
}
\caption{Model Performance on Default Thresholds}
\label{table:model_performance}
\end{table*}

\section{Ablation Study}
Table \ref{table:ablation_visualization} is an ablation study that shows the impact of chunking, humanizer label balance, and only including L1 humanizers in the train set.

\begin{table*}[h]
\setlength{\tabcolsep}{6pt}
\renewcommand{\arraystretch}{1.5}
\centering
\begin{tabular}{|l|l|c|c|c|}
\hline
\textbf{Ablation} & \textbf{Metric} & \textbf{AI (\%)} & \textbf{AI-Humanized (\%)} & \textbf{Human (\%)} \\ \hline
\multirow{3}{*}{\textbf{Final Model} } 
& TPR at 5\% FPR & $100.00 \pm 0.00$ & $98.26 \pm 0.47$ & - \\ 
& TPR at Threshold 0.5 & $100.00$ & $97.47$ & - \\ 
& FPR at Threshold 0.5 & - & - & $3.47$ \\ \hline

\multirow{3}{*}{\textbf{All-Humanizers}} 
& TPR at 5\% FPR & $100.00 \pm 0.00$ & $98.92 \pm 0.37$ & - \\ 
& TPR at Threshold 0.5 & $100.00$ & $98.93$ & - \\ 
& FPR at Threshold 0.5 & - & - & $6.00$ \\ \hline

\multirow{3}{*}{\textbf{Unbalanced}} 
& TPR at 5\% FPR & $100.00 \pm 0.00$ & $96.83 \pm 0.63$ & - \\ 
& TPR at Threshold 0.5 & $100.00$ & $95.60$ & - \\ 
& FPR at Threshold 0.5 & - & - & $3.2$ \\ \hline

\multirow{3}{*}{\textbf{Unchunked}} 
& TPR at 5\% FPR & $100.00 \pm 0.00$ & $96.69 \pm 0.66$ & - \\ 
& TPR at Threshold 0.5 & $100.00$ & $95.60$ & - \\ 
& FPR at Threshold 0.5 & - & - & $3.07$ \\ \hline

\end{tabular}
\caption{Ablation Study Results. \textbf{Descriptions:} 
\textbf{Final Model:} The final model trained using chunked samples processed by L1 Humanizers, with an equal number of humanized samples from both AI and human sources.
\textbf{All-Humanizers:} Model trained with all (L1, L2, and L3) tracked humanizers. 
\textbf{Unbalanced:} Trained without human-humanized text (all humanized samples written by AI). 
\textbf{Unchunked:} Trained on entire humanized documents without chunking into smaller segments.}
\label{table:ablation_visualization}
\end{table*}

\newpage

\section{Expanded Humanizer Audit by Source}
Table \ref{tab:text-tools} lists all humanizers and paraphrasers evaluated, with qualitative descriptions and tier rankings for each.

\begin{table*}[h]
\centering
\begin{tabular}{|p{2.5cm}|p{11cm}|c|}
\hline
\textbf{Name} & \textbf{Description} & \textbf{Tier} \\
\hline
DIPPER & Sparing and shows restraint with changes. Often most or part of a sentence is entirely unchanged. Occasionally splits sentences or adds grammar problems. & L1 \\
\hline
GPTInf & High quality text. Very few issues with spelling, punctuation, or vocabulary. & L1 \\
\hline
Grammarly & High quality text. Good varied use of punctuation. Very occasionally makes unusual edits like double quotations around a title or adding unexpected words. & L1 \\
\hline
HumanizeAI.pro & High quality text. Good grammar, advanced vocabulary, and good punctuation. & L1 \\
\hline
Quillbot & Produces flowery text but still fluent and readable. Vocabulary level is high, though slightly imprecise. & L1 \\
\hline
Semihuman AI & Good quality text. Occasionally introduces personal pronouns even when they aren't present in the source material. & L1 \\
\hline
StealthGPT & Good quality text. Output closely matches style of original text.   & L1 \\
\hline
Twixify & Overall good quality text. Occasionally misuses of words due to dictionary lookup replacements. & L1 \\
\hline
AIHumanizer.com & Generally downgrades the text from university-level to middle-school level. Lowers vocabulary level and introduces punctuation mistakes. & L2 \\
\hline
BypassGPT & Leans heavily on dictionary lookup paraphrasing. Each sentence contains the same information as a corresponding sentence in the original text. No typos or grammar errors, but occasionally the introduced words are used incorrectly or in the wrong context. & L2 \\
\hline
Stealthwriter.AI & Reduces quality of the text. There are grammar, punctuation, capitalization issues, generally one per paragraph. & L2 \\
\hline
Surfer SEO & Degrades the quality of text. Output is middle school-level writing. & L2 \\
\hline
Ghost AI & Splits every sentence into single-clause statements. Makes the output unnatural and low-quality. & L3 \\
\hline
Hix Bypass & Typically good, maybe some dictionary lookup dissonance. Occasionally there are dense pockets of nonsensical text. & L3 \\
\hline
HumanizeAI.io & Introduces fictional citations, series of question marks, and punctuation errors. The result looks like an error-ridden draft of a paper. & L3 \\
\hline
Humbot AI & Sentences are uninterpretable and random additions to the text make it unreadable. & L3 \\
\hline
Phrasly & Poor quality sentences. Much worse in some texts rather than others. & L3 \\
\hline
Undetectable AI & Poorly written text at an elementary school level. Introduces typos. & L3 \\
\hline
WriteHuman.ai & Poorly written text at an elementary school level. Occasionally includes incomplete sentences. & L3 \\
\hline
\end{tabular}
\caption{Humanizer Audit Per-Source Summaries.}
\label{tab:text-tools}
\end{table*}

\newpage


\end{document}